\PassOptionsToPackage{unicode}{hyperref}
\PassOptionsToPackage{hyphens}{url}
\documentclass[
]{article}

\usepackage{amsmath,amssymb}
\usepackage{iftex}
\ifPDFTeX
  \usepackage[T1]{fontenc}
  \usepackage[utf8]{inputenc}
  \usepackage{textcomp} 
\else 
  \usepackage{unicode-math} 
  \defaultfontfeatures{Scale=MatchLowercase}
  \defaultfontfeatures[\rmfamily]{Ligatures=TeX,Scale=1}
\fi
\usepackage{lmodern}
\ifPDFTeX\else
\fi
\IfFileExists{upquote.sty}{\usepackage{upquote}}{}
\IfFileExists{microtype.sty}{
  \usepackage[]{microtype}
  \UseMicrotypeSet[protrusion]{basicmath} 
}{}
\makeatletter
\@ifundefined{KOMAClassName}{
  \IfFileExists{parskip.sty}{%
    \usepackage{parskip}
  }{
    \setlength{\parindent}{0pt}
    \setlength{\parskip}{6pt plus 2pt minus 1pt}}
}{
  \KOMAoptions{parskip=half}}
\makeatother
\usepackage{tikz}
\usetikzlibrary{fit, positioning, shapes.geometric}
\usepackage{tikz-cd}
\usepackage{xcolor}
\usepackage[margin=1.5in]{geometry}
\usepackage{longtable,booktabs,array}
\usepackage{calc} 
\makeatletter
\patchcmd\longtable{\par}{\if@noskipsec\mbox{}\fi\par}{}{}
\makeatother

\usepackage{graphicx}
\makeatletter
\def\maxwidth{\ifdim\Gin@nat@width>\linewidth\linewidth\else\Gin@nat@width\fi}
\def\maxheight{\ifdim\Gin@nat@height>\textheight\textheight\else\Gin@nat@height\fi}
\makeatother
\setkeys{Gin}{width=\maxwidth,height=\maxheight,keepaspectratio}
\makeatletter
\def\fps@figure{htbp}
\makeatother
\usepackage{svg}
\usepackage{adjustbox}
\providecommand{\tightlist}{%
  \setlength{\itemsep}{0pt}\setlength{\parskip}{0pt}}
\makeatother
\ifLuaTeX
  \usepackage{selnolig}  
\fi
\usepackage{bookmark}
\IfFileExists{xurl.sty}{\usepackage{xurl}}{} 
\hypersetup{
  pdftitle={Reframing the Mind-Body Picture},
  pdfauthor={Ryan Williams},
  hidelinks,
  pdfcreator={LaTeX via pandoc}}

\usepackage{natbib}

\title{Reframing the Mind-Body Picture}
\makeatletter
\providecommand{\subtitle}[1]{
  \apptocmd{\@title}{\par {\large #1 \par}}{}{}
}
\makeatother
\subtitle{Applying Formal Systems to the Relationship of Mind and
Matter}
\author{ Ryan
Williams\vspace{0.05in} \\ \newline\normalsize\url{ryan@cognitivemechanics.org} }
\date{April 2024}

\begin{document}
\maketitle
\begin{abstract}
This paper aims to show that a simple framework, utilizing basic formalisms from set theory and category theory, can clarify and inform our theories of the relation between mind and matter.
\end{abstract}

I've found that theories of the mind-body relationship tend to cause
three core difficulties in my understanding:

\begin{enumerate}
\def\labelenumi{\arabic{enumi}.}
\tightlist
\item
  I'm usually left with a fuzzy picture of the relationship the author
  is trying to illustrate. When in some cases it does appear to clarify
  itself, I find myself later questioning whether the image I hold is
  actually the one intended.
\item
  I often suspect that the difference between certain systems is purely
  one of terminology, and not necessarily a disagreement in substance.
\item
  I almost never have a sense of how it could be empirically
  distinguished whether a particular theory maps to reality.
\end{enumerate}

This paper aims to make some progress against these issues by
introducing a framework for clearly expressing the ideas behind theories
of the mind-matter relationship. It consists of three parts.

In order to work with these theories on equal conceptual grounds, we
require a common philosophical foundation that allows the different
theories to be engaged with on the same terms. To address this need, the
first part of the paper outlines a conservative philosophical approach
that gives us these equal terms without presupposing theoretical
categories and relationships that the theories themselves deal with.

In the second part, we lay down formal descriptions of several specific
theories (e.g.~materialism, idealism, dualism, pansychism, etc.) on top
of that foundation. With each system resting on equal footing, we can
compare relationships between theories and pick out conceptual issues
within them. I have found the formal descriptions very helpful to
clarify my own thinking. This method of analysis offers several results
that I think demonstrate that there is fruit to be found in this
approach.

Our ultimate goal is to bring theories of the connection between mind
and matter away from the realm of \emph{a priori} argument towards
theories that can be investigated experimentally. In the third part, I
draw from my own previous work in the book \emph{Cognitive
Mechanics},\footnote{\citealt{williams2022cognitive}} which sets out
empirically-motivated mathematical properties\footnote{The work is
  consistent with, but not motivated by \citealt{Lee2022}.} we might look for
in material systems that correlate with mental processes.\footnote{Throughout
  the paper I will use the terms ``physical'' and ``material''
  equivalently; as well as ``things'', ``objects'' and ``entities''.
  Unless a specific definition of a term is given, it should be taken
  that a common-sense meaning is sufficient.} We explore the theories
from Part II, implemented in the context of a more specific theory. This
process arrives at a set of critical \emph{equivalence properties} which
enable a novel categorization of the theories. Part III concludes with an example
that illustrates a method for connecting theories expressed within the
framework with experimental observations.

\subsection{Part I: Framework}\label{part-i-framework}

The goal of our framework is not to produce an unassailable metaphysical
position. Instead, we aim only to build a conservative, pragmatic
starting point that allows us to examine specific theories of the
relationship between mental and physical phenomena without undue \emph{a
priori} bias.

\subsubsection{Assumptions}\label{assumptions}

We will utilize only the following two assumptions to build our
theories:

\begin{enumerate}
\def\labelenumi{\arabic{enumi}.}
\tightlist
\item
  \emph{Object Pluralism:} We will admit the statement of theories in
  terms of distinct categories, entities, properties, and relations
  between them (more generally we'll term these \emph{elements}). These
  elements will be expressed formally as sets, graphs, hypergraphs, and
  categories.
\item
  \emph{Structural Agnosticism}: We will make no assumptions about the
  structure of the elements within our framework. Our \emph{theories}
  are allowed to draw distinctions, but they must define their elements
  and specify how they relate to each other.
\end{enumerate}

Our models below will be built from (and subject to) these two features.

\subsubsection{Formal Descriptions}\label{formal-descriptions}

Consistent with our assumption of object pluralism, we will utilize
formal categories, sets, graphs, and hypergraphs to represent different
theories of the mind-body relationship. For instance, discernible
elements may be expressed as nodes of a graph or members of a set;
relationships between them could be stated as hyperedges that connect
them, or by membership of the same set or category.

We will use the simplest formalism for each representation that conveys
the underlying ideas, sometimes at the expense of notational consistency
between models. For example, in one model the material elements may be
expressed as a set where no further specification is needed, while it
may be expressed as a hypergraph or formal category in others.

Drawing from our structural agnosticism, we do not have any prejudice
whether these elements are termed ``mental'' or ``physical'', or what
the elements themselves \emph{are}. It is left to the theories to define
their categories and relationships, and to justify their structure via
explanatory contributions.

Since our primary concern is in the relationship between the elements,
it is largely left unspecified exactly which physical and mental objects
and relations are being represented. These mental and physical types
should be understood broadly.

\subsection{Part II: Comparing Models of the
World}\label{part-ii-comparing-models-of-the-world}

We will now take our simple foundation and use it as the medium for
expressing the various theories of the relationship between mind and
matter. This exercise will not only allow us to more clearly express the
similarities and differences between theories for comparison, but will
also allow us to potentially extract empirical features that would
provide evidence for the utility of one theory versus others.

\subsubsection{Solipsism}\label{solipsism}

While our principle of object pluralism may strike with a note of
metaphysical realism (the idea that things exist independently of being
thought or experienced), it actually does not imply as much. For
instance, you could define a theory in which the set \(U\) of all
elements of the world was identical to the set of mental elements \(I\).
This is perfectly accepted within our framework. \[ U = I \] If the
elements of \(I\) are undifferentiated and make up a single subject,
this is the statement of solipsism, the position that the world consists
of only one mind, and does not contain material elements or other minds.
Solipsism is a form of idealism, which we will address in more detail
below.

\begin{figure}
    \vspace*{3em}
    \hspace*{.55in}
    \adjustbox{width=\textwidth,height=0.25\textheight,keepaspectratio}{\includegraphics{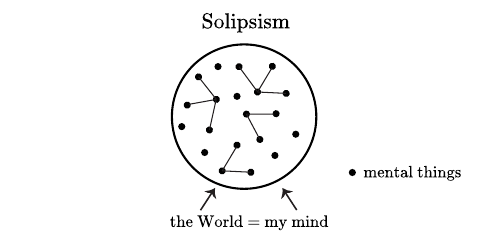}}
    \caption{Solipsism is the statement that the world is made up of only
mental things, and that the collection of all mental things is a single
mind.}
\end{figure}

\pagebreak

\subsubsection{Materialism}\label{materialism}

The dominant current stance in the sciences and philosophy is one of
materialism (or physicalism more broadly). Materialism can be viewed as
the statement that the objects of the world are material, while certain
relations between objects pick out what we normally consider to be
minds. The main stance within materialism is emergentism, in which the
mind emerges from physical states or interactions.

Materialism encounters the objection that the phenomena of consciousness
seem nowhere to be found in our physical descriptions of the world. The
result is an ``explanatory gap'' in which the physical constituents
don't apparently have any necessary connection to mental
phenomena.\footnote{\citealt{levine1983materialism}} Opponents point out that the approach
still doesn't explain what the mind \emph{is}.

We say that materialism takes the world to be a hypergraph
\(U = (M, R)\) where the set of vertices \(M\) is equivalent to the
material objects and \(R\) is the set of hyperedges that represent
relations between material systems, i.e.~subsets of \(M\).

We then take the mental elements of this framework \(I\) to exist as a
subset of the relations \(R\) that meet some predicate \(Q\), that is
\(I = \{ r \in R : Q(r) \}\). \(Q\) determines which relations are those
of mind and which others are not. Distinguishing between different
material theories of mind, then, amounts to distinguishing between
different versions of the condition \(Q\).
\[U = (M, R), \ I = \{ r \in R : Q(r) \}\]

\subsubsection{Panpsychism}\label{panpsychism}

Panpsychism is a flavor of materialism in which fundamental constituents
of matter have a type of proto-consciousness, combinations of which
result in higher forms of consciousness, such as that exhibited in
humans. It is subtle to distinguish from emergentism.

The materialist system outlined above can represent \emph{both}
emergentism and panpsychism. The difference is that in panpsychism,
\(Q(r)\) is true for all \(r\); which is equivalent to the stipulation
that \(R = I\), i.e.~that all material systems are in some way mental,
in addition to being physical. In practice, many forms of panpsychism
will include a concept of the level of mentality or consciousness of a
system varying on a smooth gradient, as with Integrated Information
Theory in the following section, which is worthy of further explication.

If one takes the viewpoint that we cannot distinguish the essence of
either physical or mental things without defining their relationships
and causal properties, it could be argued that this distinction amounts
to simply an ill-defined notion of what it is to be mental without any
empirical justification. An opponent could charge that the panpsychist
would still need a \(Q'\) to pick out a subset \(I' \subset R\) of the
entities we would normally consider to be mental; in that case you'd be
back to an equivalent model (i.e.~the emergentist \(Q = Q'\) and the
emergentist \(I = I'\)).

From my own perspective, it doesn't seem on its face that there is
anything that precludes ``physical'' stuff from having qualitative
properties, so long as it isn't explicitly defined not to have any such
properties. That would seem to undermine the evident motivation behind
the panpsychist move. If even conscious systems are often unconscious
(e.g.~when we sleep, are under anesthesia, etc.), it could seem a
stretch to postulate that seemingly unconscious systems have conscious
properties.

The defense that these are ``proto-conscious'' properties could add
weight to the suspicion that these differences may be semantical or
ill-defined. On the other hand, it seems that the emergentist is already
committed to some form of latent mental properties in the material. Our
system shows deep connections here, and the distinctions may largely be
of emphasis and terminology.\footnote{This conclusion is parallel to
  those found in \citealt{strawson2006realistic}, in which he took the position that
  materialists \emph{must} hold the panpsychist view to be
  intellectually consistent.}

\emph{Neutral monism}, outlined below, is another theory that takes its
fundamental constituents to have both mental and physical aspects. It is
difficult to draw distinctions between neutral monism and panpsychism
without the clarification that neutral monism is modeled very
differently from \(R = I\). We will see in Part III that neutral monism
has a fundamental distinction by a property we call \emph{mutual
non-equivalence}.

The sharpness and ease of these comparisons demonstrate the potency of
our framework. It clarifies the statements of each system in a way that
allows us to precisely distinguish between them (or potentially indicate
their similarity, as shown above).

\subsubsection{Integrated Information
Theory}\label{integrated-information-theory}

Integrated Information Theory is a materialist theory that contains a
method for calculating a value \(\Phi\) that represents the level of
consciousness of an abstract system based on its cause-effect
structure.\footnote{\citealt{oizumi2014phenomenology}}

There are practical hurdles to calculating \(\Phi\) in actual physical
systems due to reasons of computational tractability, because it relies
on counterfactual properties of the physical system, and because it is
not specified in the theory what level of physical description must be
used to calculate the value.\footnote{This ambiguity in level of
  description is actually a common feature with \emph{Cognitive
  Mechanics}, explored in Part III, although they are arrived at in
  different ways.}

Nevertheless, we can express the theory succinctly in our framework. Not
surprisingly, as a materialist theory, its definition involves a
specification of the condition \(Q\). IIT takes \(Q\) as the statement
that \(\Phi\) is greater than some \(\tau\), that is
\[Q(r) = \Phi_r > \tau.\] In other words, the set \(I\) of mental
entities contains all physical systems with a value of \(\Phi\) over the
threshold \(\tau\). It may be natural to suppose \(\tau = 0\), which
would mean that a very large subset of \(R\) would be considered to have
some form of mental instantiation.

\subsubsection{Illusionism}\label{illusionism}

Among materialists are also so-called illusionists (also
eliminativists), those who claim mental phenomena are ``illusions''. In
practice, this position often looks exactly like our system of
materialism above, in which there still exists some criteria \(Q\) that
delineates ``mental'' constructions from the physical. If our view is
correct, our framework is highlighting that this is merely a semantical
difference and not one of substance.

\begin{figure}
\hspace*{1in}
\adjustbox{width=\textwidth,height=0.2\textheight,keepaspectratio}{\includegraphics{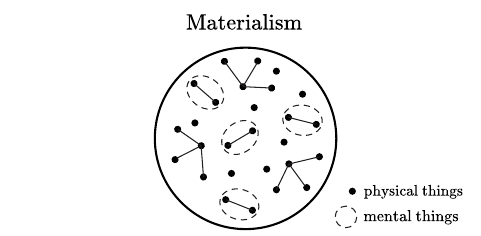}}
\caption{Materialism takes the things of the world to be material, while
mental things arise out of specific patterns of material elements.}
\end{figure}

\begin{figure}
    \hspace*{1in}
\adjustbox{width=\textwidth,height=0.2\textheight,keepaspectratio}{\includegraphics{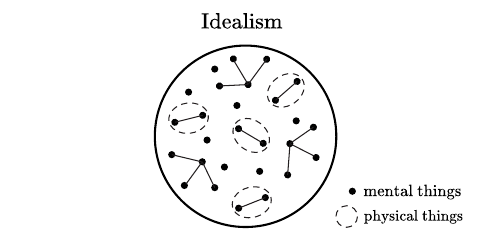}}
\caption{Idealism instead considers the things of the world to be
mental, while physical things arise out of specific patterns of mental
elements. Note the mirror relationship with the diagram of materialism.}
\end{figure}

\begin{figure}
\hspace*{1in}
\adjustbox{width=\textwidth,height=0.2\textheight,keepaspectratio}{\includegraphics{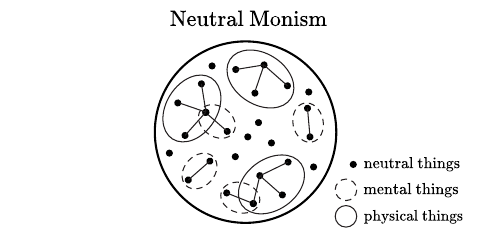}}
\caption{Neutral monism states that both physical and mental things are
made up of the same underlying neutral things, which are neither wholly
physical nor wholly mental. Notice that whether a system is categorized
as physical or mental is determined by its relationships, as opposed to
its substance.}
\end{figure}

There are those who take a more hard-lined illusionist view in which the
illusion is total: there exist no criteria \(Q\) to discriminate mental
things from the physical. We can state this formally if \(U\) is the set
of all elements of the world, and \(M\) is the set of all physical
elements, then \(U = M\) without any further discrimination.

Their perspective is dual to that of solipsism in the sense that where
solipsism denies that anything outside a single mind exists,
illusionists deny that anything outside an undifferentiated material
system exists.

\subsubsection{Idealism}\label{idealism}

From the other side, many idealists recognize the same issues as the
materialists and take the problem in the opposite direction. Seeing no
fruit in the materialist approach, they attempt to ground the physical
as emergent from the mental. They see the puzzling revelations of modern
physics and the trajectory towards fundamental elements that seem to
becoming less ``physical'' with every discovery. They say that, given
that we don't know what the physical stuff \emph{is}, we'll start with
the mental stuff as fundamental.

As a theory, idealism is the inverse of materialism. The world consists
of a set of mental entities, and the material world is derived from the
relations between the mental entities.

Formally, idealism is the statement that the world is a hypergraph
\(U = (I, R)\) where the set of objects \(I\) are mental, while the
material objects \(M\) are constituted of a subset of the relations
\(R\) that meet criteria \(Q\),
\[U = (I, R), \ M = \{ r \in R : Q(r) \}.\] Compare with the above
definition of materialism. That these ideas should be dual to each other
in this way is illuminating, but once stated should not be surprising.
Both materialism and idealism carry the same reduction of one category
into another.

\subsubsection{Neutral Monism}\label{neutral-monism}

In the mid-19th century and into the early 20th, great minds from a
variety of fields including Ernst Mach\footnote{\citealt{mach2010analysis}}
(physicist), William James\footnote{\citealt{james1975pragmatism}} (psychologist), and
Bertrand Russell\footnote{\citealt{russell1921analysis}} (mathematician) adopted
slightly differing but similar stances that are neither the physicalist
nor the idealist perspective. Their \emph{neutral monism} instead claims
that all things of the world are of the same fundamental type, and that
the physical and mental aspects of the world manifest in different
perspectives, roles, relationships, etc. between these elements of the
same fundamental nature. The details of each is slightly different
depending on the author.

This neutral perspective has the advantage that it does not propose two
separate categories of phenomena, but that both are aspects of a single
underlying ``neutral'' substance that is neither wholly mental nor
wholly physical. Each of the formulations has its own peculiar details
about the nature of this single underlying substance.

Neutral monists tend to place the emphasis on relations between elements
as opposed to \emph{what} the elements are.

They take the world as a hypergraph \(U = (N,R)\), where \(N\) is the
set of neutral objects that underlie reality; and \(R\) is a set of
hyperedges that represent systems of the neutral objects, with the set
of mental objects \(I \subset R\) and the set of material objects
\(M \subset R\). There are two predicates, \(Q_I\) and \(Q_M\) which
determine whether a hyperedge is a member of \(I\) or \(M\),
respectively. Then,
\[U = (N, R), \ I = \{ r \in R : Q_I(r) \}, \ M = \{ r \in R : Q_M(r) \}.\]
It is likely apparent to the reader the close association this statement
has with that of both materialism and idealism. Instead of the material
arising from the mental or vice versa, the mental and material both
arise from the underlying neutral substance.

\subsubsection{Russellian Monism}\label{russellian-monism}

Russellian Monism is a type of neutral monism proposed by Bertrand
Russell that has a specific structure. In Russell's monism, the world is
a set of neutral \emph{particulars}. These particulars are in causal
connection with other particulars.

A physical object is a \emph{causal complex}, the collection of all
effects a given set of particulars has at a point in time. A
\emph{perspective} is the collection of effects of all the particulars
in a given place, and corresponds in some cases with a subject of
experience.

We can state that the world is an \emph{ordered} graph \(U = (N,R)\)
where each \(r \in R\) is an ordered pair \((v, w)\), and \(v\) is
understood to be the cause of the effect \(w\).

If \(V(x) \subset R\) is the set of relations where \(v = x\), and
\(W(y) \subset R\) is the set of relations where \(w = y\), then this
gives
\[ M = \{V(x) : x \in N\}, \ I = \{W(y) : y \in N\}.\]
It can be seen that \(I \subset 2^{R}\) and \(M \subset 2^R\), where \(2^R\) denotes the power set of \(R\).

\begin{figure}
\vspace*{.25in}
\hspace*{.85in}
\adjustbox{width=\textwidth,height=0.22\textheight,keepaspectratio}{\includegraphics{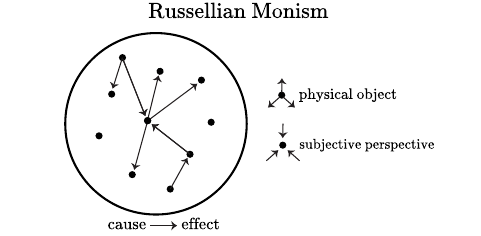}}
\caption{Russellian monism presents physical and mental elements as
different views of the same underlying neutral substance, where the view
of a physical object is the collection of its effects, while a
subjective perspective is the collection of effects at a given place.
Note that the same node may consititute a physical object or a
subjective perspective depending on the relationships being examined.}
\end{figure}

\subsubsection{Substance Dualism}\label{substance-dualism}

Modern philosophy is often pegged to begin with Descartes, whose
substance dualism gives a somewhat intuitive notion of the mind and
matter being nearly completely separate realms.\footnote{\citealt{descartes1641meditations}} This view can be found as far back as Plato's separate
treatment of body and soul.\footnote{\citealt{plato1993phaedo}}

The dualist approach has largely fallen out of favor due to its supposed
lack of parsimony in that it proposes independent categories of
substance. I find this objection a bit odd, because it is usually
replaced by an account that still contains multiple ontological
categories, even if one is subsumed within another.

It is often additionally argued that it simply places the mind in the
role of an unexplained homunculus. That may be the case, but I haven't
yet encountered a theory that actually gives an account of what physical
\emph{or} mental substance actually \emph{is}. There seems reason to
believe that the most useful theories we can construct are those of
relationships within and between entities and categories, without
necessarily specifying what they \emph{are} fundamentally.

From our perspective, a simple statement of there being two categories
isn't enough to engage with. A well-formed theory \emph{must} specify
the relationship between any categories it puts forward.

In the naive form, dualism would posit two entirely separate categories
with no connection between them. Formally, we could state that the
intersection of \(M\) and \(I\) has no members \(M \cap I = \emptyset\),
i.e.~there is no overlap between the mental and the physical. This is an
admissible theory in our framework, but it comes with the consequence
that neither category can have a causal connection with the other.

I don't think the intent of most dualists is to say that there is no
connection whatsoever between the mental and physical world. Once
connections begin to be drawn between the mental and the physical, the
picture becomes more interesting. A formal illustration of these more
interesting forms of dualism will be discussed in Part III.

\begin{figure}
    \vspace{1em}
    \hspace*{.65in}
    \adjustbox{width=\textwidth,height=0.25\textheight,keepaspectratio}{\includegraphics{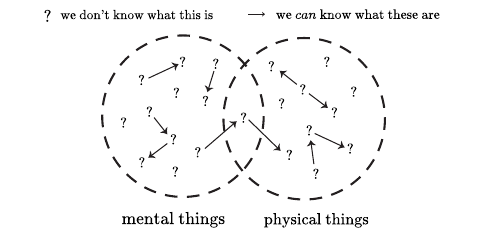}}
\caption{We are equally uncertain of the fundamental substance of both
mental and physical things. But we \emph{are} able to discover
relationships between the objects of our experience.}
\end{figure}

\subsection{Part III: Implementing Specific
Theories}\label{part-iii-implementing-specific-theories}

The methods we've been using were previously put to use in my book
\emph{Cognitive Mechanics}, though it was not formulated in the specific
terms we have used here. The book relates the behavior of mental systems
to physical systems by defining a set of empirically-motivated
operations, each of which has a differently-instantiated---but
equivalent---form in the mental system and the physical system.

To be clear, I believe this system deserves skepticism, as any other.
The broad idea of the book is to start from basic mental capabilities we
can know via immediate exercise that we possess (such as the
construction of new concepts from existing ones), and to turn the
function of those capabilities into precise mathematical descriptions of
behavioral patterns we could observe experimentally in biological
systems. There are myriad ways to misstep in this terrain.

The formalism I use here will be slightly different from the one that
appears in the book, but the notions are the same. I will not give a
full outline of all operations explored in the book, but will instead
give a flavor of the idea by way of a single operation \(C\).

Although the following part is concentrated on the ideas outlined in
\emph{Cognitive Mechanics}, our framework can and should be used to
explore other specific empirical theories, and their potential
connections with those outlined in Part II.

Due to our framework, I have come to realize more about the nature of
the system I will describe below, and have found that it elucidates key
properties of the main types of theory from Part II: materialism,
idealism, neutral monism, and dualism.

\subsubsection{Mental Category}\label{mental-category}

First, we define a category \(\mathcal{I}\) (in the category theoretic
sense) that will represent our mental system. The objects of
\(\mathcal{I}\) are referred to as \emph{concepts}. These concepts track
closely with our informal notion of what a concept is as a mental
abstraction of a category of entities. The set of all concepts will be
denoted \(X\).

We take an operation \(C\), which represents a mental operation we call
\emph{composition}. It is the ability to take two existing concepts and
form a new one, wholly in your mind.

For instance, I may tell you to imagine a purple hexagon, which your
mind can construct, even if you have only ever seen various purple
objects (but never a hexagonal one) and various hexagons (but never a
purple one).

Operation \(C\) in category \(\mathcal{I}\) is a morphism
\(C : X \times X \to X\). In other words, operation \(C\) takes two
concepts and produces a new concept. Operation \(C\) is defined
\[ C(a, b) = d, \text{ where } a , b, d \in X, \] and where \(a\) and
\(b\) are called \emph{components} of the resulting concept \(d\).

\subsubsection{Material Category}\label{material-category}

Our second category \(\mathcal{M}\) represents the material system. The
objects of \(\mathcal{M}\) are referred to as \emph{manifestations}. The
set of all manifestations is denoted \(Y\).

\begin{figure}
    \hspace*{1.05in}
    \adjustbox{width=\textwidth,height=0.2\textheight,keepaspectratio}{\includegraphics{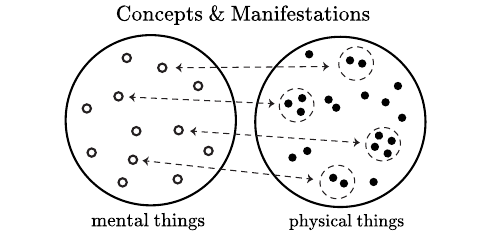}}
\caption{In \emph{Cognitive Mechanics}, mental entities called
\emph{concepts} map to sets of their correlated \emph{manifestations} in
the material world.}
\end{figure}

\begin{figure}
    \hspace*{1.05in}
    \adjustbox{width=\textwidth,height=0.2\textheight,keepaspectratio}{\includegraphics{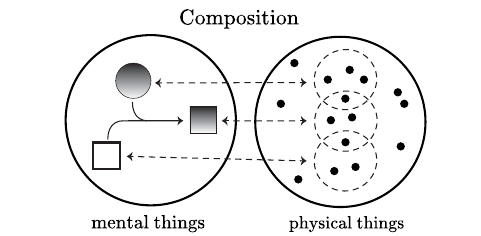}}
\caption{Operation \(C\), \emph{compose}, takes two concepts and creates
a new one from them. Physically, this equates to the resulting concept
sharing \emph{some} of its manifestations with both of its components.}
\end{figure}

\begin{figure}
    \hspace*{1.05in}
    \adjustbox{width=\textwidth,height=0.2\textheight,keepaspectratio}{\includegraphics{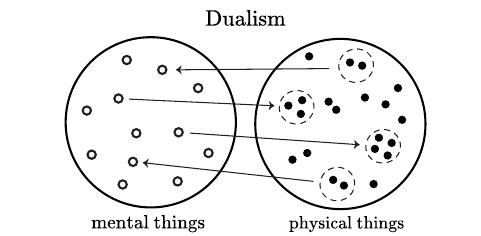}}
\caption{This diagram shows an interaction dualism that has sparse
causal connections from physical to mental and vice versa. Notice that
there are mental things which have no physical connection \emph{and}
physical things which have no mental connection; this is an illustration
of the \emph{mutual non-equivalence property}.}
\end{figure}

Manifestations refer to entities of the physical state of the system
without further definition. It is left open whether these entities are
molecules, neurons, cortical columns, neural assemblies, etc., or some
other patterns of physical state.

The idea is to find which physical level can account for operation
\(C\), by looking for mathematical properties in the behavior of the
material system we will find below.

Let us imagine a functor (i.e.~a transformation between categories)
\(F\), where \(F\) maps objects and operations from the mental category
\(\mathcal{I}\) to the material category \(\mathcal{M}\), that is
\(F : \mathcal{I} \to \mathcal{M}\).

Each concept \(x \in X\) of \(\mathcal{I}\) is represented by a set of
manifestations \(Z \subseteq Y\) in \(\mathcal{M}\), \(F(x) = Z\). Which
is to say, each concept has a set of material correlates in
\(\mathcal{M}\).

Each morphism (i.e.~operation) \(O\) in \(\mathcal{I}\) has an
equivalent morphism \(F(O)\) in \(\mathcal{M}\) that describes how the
physical manifestation of the mental process is instantiated. Meaning
each mental operation has an equivalent material operation on the
correlates of the concepts it operates on.

In the case of our operation \(C\),
\(F(C) : F(X) \times F(X) \to F(X)\). Within the category
\(\mathcal{M}\), \(F(C)\) is defined:
\[ F(C)(A, B) = S_A \cup S_B \cup S_u, \text{ where } A, B, S_u \subseteq Y, \ S_A \subseteq A, \ S_B \subseteq B. \]

In other words, the resulting concept has a manifestation that shares
properties with the manifestations of both of its components, \(a\) and
\(b\), along with an indeterminate set of other manifestations \(S_u\).
The new concept is made up of some combination of a \emph{subset} of
each of the sets of manifestations \(A\) and \(B\).

\subsubsection{Discriminating Theories by Equivalence
Properties}\label{discriminating-theories-by-equivalence-properties}

If we take a step back, we will notice that this general approach above
does not necessarily give primacy to the category \(\mathcal{I}\), nor
to the category \(\mathcal{M}\). Let us also consider a functor \(G\)
that goes from \(\mathcal{M}\) back to \(\mathcal{I}\), that is
\(G : \mathcal{M} \to \mathcal{I}\).

In the book, there are inverse mappings \(M\) and \(M^{-1}\), an
isomorphism between the two categories. Ontologically, this isomorphism
would seem to indicate that the two aspects are wholly equivalent.
Interestingly, that was not necessarily my intention, but our framework
clarifies the situation. There could be empirical reasons to doubt this.

If you believed for empirical or theoretical reasons that there are
material entities or phenomena that have no mental equivalent (i.e.~that
\(F(\mathcal{I})\) is non-surjective into \(\mathcal{M}\)), that could
be motivation to reformulate the theory as a materialistic one, with the
result that \((F \circ G) (\mathcal{M}) \neq \mathcal{M}\), while still
allowing \((G \circ F) (\mathcal{I}) = \mathcal{I}\). For example, there
seems \emph{prima facie} evidence that the material world exists in some
consistent state, without regard to whether it is being observed, and
that there exist material systems with no corresponding mental
phenomena. We will call this property \emph{partial equivalence}.

In the opposite direction, if you had empirical evidence that there are
mental entities or phenomena that had no material equivalent, it would
indicate \((G \circ F) (\mathcal{I})\neq \mathcal{I}\), but still allow
\((F \circ G) (\mathcal{M}) = \mathcal{M}\). This is the idealist
stance. Defenders of this position would point to phenomenological
properties or qualia, which don't seem to have a necessary connection to
the material regularities of the world.

However, neutral monists and dualists can say \emph{both} that there are
mental entities or phenomena with no material equivalent \emph{and} that
there are material entities or phenomena with no mental equivalent,
i.e.~that \((G \circ F) (\mathcal{I})\neq \mathcal{I}\) and
\((F \circ G) (\mathcal{M}) \neq \mathcal{M}\). This option is not
available in materialism and idealism, which both presuppose that one
category is subsumed within the other.

We will call this the \emph{mutual non-equivalence property}. In
addition to the motivations given by independence of material states and
phenomenological properties of experience, the mutual non-equivalence
property matches well to common intuitions about incongruencies between
physical and mental phenomena that motivate the conception of the hard
problem in the first place.\footnote{\citealt{chalmers1995facing}}

Neutral monists implement this idea via a new construction in which
there is one neutral category \(\mathcal{U}\) which has functors
\(F' : \mathcal{U} \to \mathcal{M}\) and
\(G' : \mathcal{U} \to \mathcal{I}\). This is a system in which the
properties of the physical and mental derive from an underlying neutral
substance in \(\mathcal{U}\). A monist would interpret that
\(\mathcal{U}\) is the underlying reality, while both \(\mathcal{I}\)
and \(\mathcal{M}\) are different manifestations of the same underlying
substance.
\[
  \begin{tikzcd}
    & {\mathcal{M}} \arrow[dd,shift left=.7ex, "G"] \\
    {\mathcal{U}} \arrow[ru, "F'"] \arrow[rd, "G'" swap]   &     \\
    & {\mathcal{I}} \arrow[uu, shift left=.7ex, "F"]
  \end{tikzcd}
\]
Importantly, neutral monism permits the mutual non-equivalence property:
note that there is no necessity that there is an isomorphism between
\(\mathcal{I}\) and \(\mathcal{M}\), nor that there is any functor \(H\)
which is surjective from either \(\mathcal{I}\) or \(\mathcal{M}\) onto
\(\mathcal{U}\). However, neutral monism also permits partial
equivalence, which it would be reasonably argued would be described more
sharply by a fully materialistic or idealistic theory.

Dualists who recognize interactions between the physical and mental
categories would instead interpret \(F\) and \(G\) as sparse causal
links rather than mere mapping of equivalent structures. Interestingly,
while dualism is generally dismissed for reasons of parsimony, there are
reasonable arguments that it is formally the more parsimonious system of
the two that hold the mutual non-equivalence property. We shouldn't lose
sight that the sparse causal links between the categories need to be
demonstrated empirically, which is clearly a significant hurdle. But
we've found that the standard arguments against dualism for reasons of
parsimony don't seem to stand to scrutiny in our framework.

It seems that every system permits \emph{full equivalence}; that is:
\[(F \circ G) (\mathcal{M}) = \mathcal{M}, \ (G \circ F) (\mathcal{I}) = \mathcal{I}.\]
If the systems are fully equivalent, a case could be made that
emergentist or idealist claims of one system being more fundamental than
the other are spurious; though it seems compatible with panpsychist,
neutral monist, and dualist views.

In the panpsychist case, \(F\) and \(G\) could be seen as the identity
functor \(1_{\mathcal{M}}\), potentially with the removal of the
category \(\mathcal{I}\) altogether. This may seem like a parsimonious
move, but the resulting system resembles an eliminitivist system such as
\(U = M\). The question then becomes whether the system can account for
different properties of mental and physical phenomena, and seems to make
the claim of mental properties within the physical superfluous.

For neutral monists, full equivalence would seem to indicate an
additional condition of isomorphism between \(\mathcal{I}\) and
\(\mathcal{M}\). This would be something like what is known as a
\emph{dual aspect monism}, in which the mental and physical correspond
but are irreducible to each other. In the case of Russellian monism, it would indicate the condition
of a symmetric graph \((N, R)\) where every edge \((v, w) \in R\) has an
associated \((w, v)\).

For dualists, this equivalence would also indicate an isomorphism
between \(\mathcal{I}\) and \(\mathcal{M}\). This may be interpretted as
a \emph{property dualism}, in which the mental and physical are distinct
but correlated properties, but not different substances.

\begin{figure}[h]
  \includegraphics{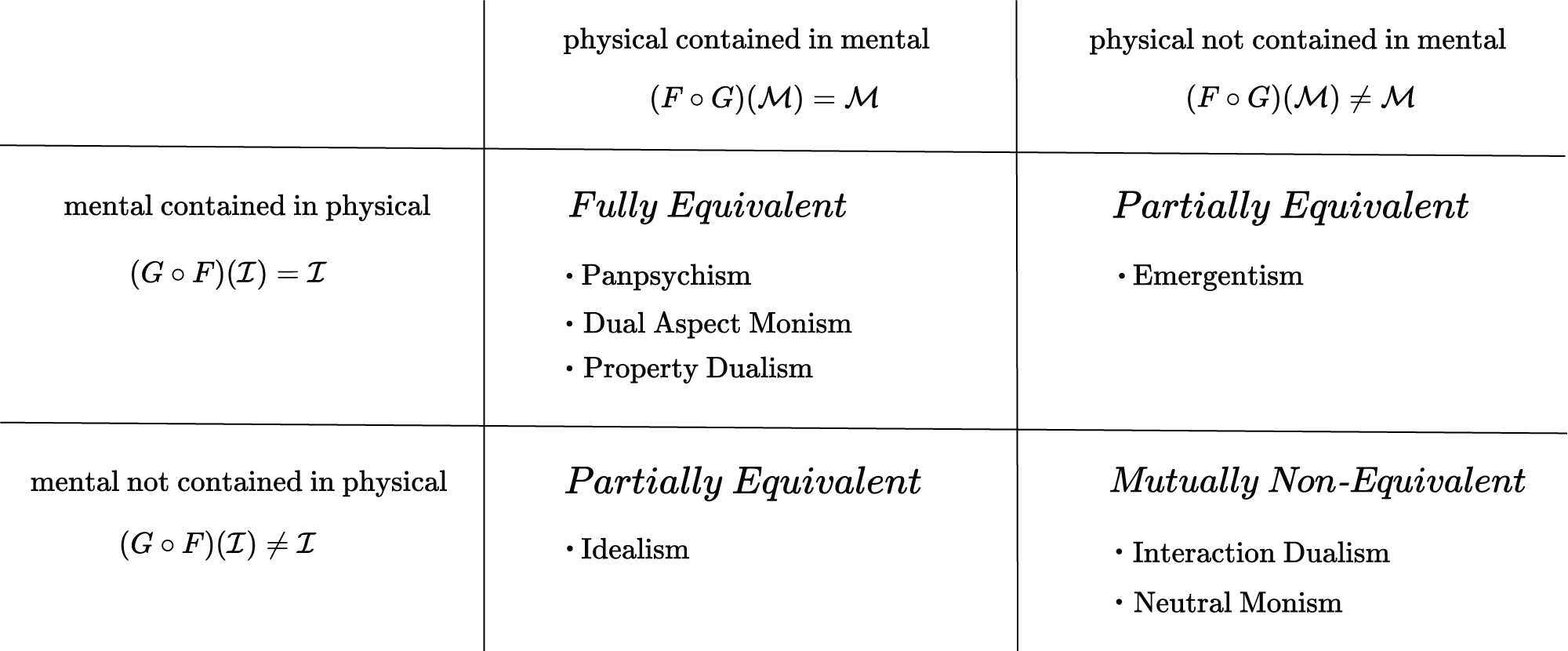}
  \caption{This diagram shows a comparison of mind-matter theories
  according to equivalence properties.}
  \label{diag}
\end{figure}

\subsubsection{Drawing Out Empirical
Consequences}\label{drawing-out-empirical-consequences}

In this section, we illustrate a process for deriving empirical
consequences of the system outlined in \emph{Cognitive Mechanics}. The
purpose of this section is not necessarily to make a compelling case for
a specific theory, but rather to demonstrate a methodology for gathering
experimental predictions from theories expressed within our framework
that we could compare with observed data. It is put forward as a sketch
rather than a finished proposal. There are simplifications in the
formulation below, specifically of the parameters \(\alpha\) and
\(\beta\), that will likely need to be further expounded.

Within existing neuroscience, there is a vast and mature field in
gathering and interpreting brain imaging data. There is substantial work
to be done just to compare with existing results. If the specific ideas
of \emph{Cognitive Mechanics} do find use---which is by no means a
foregone conclusion---it is possible that they could be more valuable in
interpreting existing results as mental processes than proposing novel
physical descriptions. Though I think that the compositional nature of
operations in the system could provide interesting predictions.

In this simplified example, I am choosing neurons as instances of the
manifestations. We would ideally want to examine multiple physical
levels, including higher-level structures like cell assemblies, cortical
columns, brain regions, and potentially lower-level structures such as
molecules. With that ability difficult at this point in time, we may be
able to utilize techniques to infer activity at other levels from brain
imaging voxel data.\footnote{Interesting work has been done already in
  \citealt{Nishimoto2011}, \citealt{Kriegeskorte2010},
  \citealt{Naselaris2011}.} Assuming correlations between levels of
physical instantiation, it's possible there are results to be found
within the raw data as well.

There is work to be done to adapt the statistical prediction below to
generate predictive datasets.\footnote{\citealt{mitchell2008predicting}
  demonstrates a methodology for predicting fMRI states for specific
  words that could be drawn upon.} There are also representational
subtleties that could motivate elaborations to the model.\footnote{For
  example, the binding problem explored in \citealt{9528907}. The paper also demonstrates ways of restating sparse
  representations as dense vectors that could prove to be useful.}

Caveats out of the way, the idea behind the formulation of \(C\) above
is to give a specific physical relationship that one can look to find in
the material correlates of operation \(C\), manifested as \(F(C)\).

Imagine that we take the brain to be a simple collection of neurons
\(Y = \{n_1, n_2, ..., n_N\}\). An active state of some concept \(x\) is
some subset \(F(x) \subset Y\) of those neurons that are active at the
time that concept \(x\) is considered to be experienced. The table below
shows a list of concepts, along with each concept's active neurons.

Note that each composed concept shares active neurons with each of its
components. For instance, the \emph{green circle} shares neurons \(n_1\)
and \(n_2\) with the concept \emph{circle}, along with \(n_6\) and
\(n_9\) with the concept \emph{green}.

\begin{longtable}[]{@{}ll@{}}
\toprule\noalign{}
Concept & Active Neurons \\
\midrule\noalign{}
\endhead
\bottomrule\noalign{}
\endlastfoot
square & \(n_1, n_2, n_3, ...\) \\
circle & \(n_1, n_2, n_4, ...\) \\
ellipse & \(n_1, n_4, n_5, ...\) \\
red & \(n_6, n_7, n_8, ...\) \\
green & \(n_6, n_7, n_9, ...\) \\
blue & \(n_6, n_9, n_{10}, ...\) \\
red square & \(n_1, n_3, n_7, n_{8}, ...\) \\
green circle & \(n_1, n_2, n_6, n_{9}, ...\) \\
blue ellipse & \(n_4, n_5, n_6, n_{10}, ...\) \\
\end{longtable}

We can use these formal properties to deduce statistical predictions
about the physical system. Given the concept \(x\), the active set of
neurons is \(F(x)\).

We will take \(p(n_i|x)\) as the probability that neuron \(n_i\) is
active in concept \(x\) and \(p(n_i|x,y)\) to be the probability that
\(x\) and \(y\) both share the active neuron \(n_i\). If \(|F(x)|\) is
the quantity of active neurons in concept \(x\),
\[ p(n_i|x) = \frac{|F(x)|}{N}, \] and \(p(n_i|x,y)\) can be stated as
\[ p(n_i|x,y) = \frac{|F(x) \cap F(y)|}{N}. \]

We can state the probability that \(n_i\) is active in \emph{either}
\(v\) or \(w\) as
\[ p(n_i|v \vee w) = p(n_i|v) + p(n_i|w) - p(n_i|v, w). \]

If \(p(n_i|x)\) and \(p(n_i|y)\) are independent,
\[ p(n_i|x,y) = p(n_i|x) \cdot p(n_i|y). \]

We will take it that the quantity of active neurons \(\alpha = |F(x)|\)
in a conceptual state \(x\); then \( p(n_i|x) = \alpha / N \) and
\[ p(n_i|v \vee w) = \frac{2\alpha}{N} - \frac{\alpha^2}{N^2}. \]

Given that the number of active neurons for a concept \(x\) selected
from each component \(u\) is \(\beta = |F(x) \cap F(u)|\), we can now
state \(p(n_i|x;v,w)\), the probability that \(n_i\) is active in the
concept \(x\) composed of concepts \(v\) and \(w\): \[ 
p(n_i|x;v,w) = \begin{cases}
                 \frac{ s_{vw} }{ N_{vw} } & \text{when } n_i \in F(v) \cup F(w), \\
                 \frac{ \alpha - s_{vw} }{ N - N_{vw} }  & \text{otherwise,}
\end{cases}
\]
where
$N_{vw} = 2\alpha - \frac{\alpha^2}{N}$, the expected number of neurons in $F(v) \cup F(w)$, and
$s_{vw} = 2\beta - \frac{\beta^2}{N}$ is the expected number of neurons sampled from $F(v) \cup F(w)$, taking into account
expected neurons in common between samples.

\subsection{Results}\label{results}

In the course of this article, several results were arrived at as a
direct product of the framework we've introduced.

\begin{enumerate}
\def\labelenumi{\arabic{enumi}.}
\tightlist
\item
  Materialist and idealist theories are formally dual to each other in
  an interesting way.
\item
  The distinction between different materialist and idealist theories
  can be formulated as alternate versions of a single predicate \(Q\)
  over the set of material and mental systems, respectively; likewise
  with neutral monism, except with two predicates \(Q_I\) and \(Q_M\)
  over a set of relations between neutral entities.
\item
  There are deep connections between panpsychist and emergentist models,
  and sharper distinctions than there may seem \emph{prima facie}
  between panpsychism and neutral monism.
\item
  Many forms of illusionism are restatements of emergent materialism.
\item
  Essential distinctions of various theories were stated in terms of
  equivalence properties, derived from functors on the material category
  \(\mathcal{M}\) and the mental category \(\mathcal{I}\). \emph{Full
  equivalence}, \emph{partial equivalence}, and \emph{mutual
  non-equivalence} serve as powerful tools for comparison between
  theories with fundamentally different ontologies.
\item
  Dualism should not be dismissed \emph{a priori} for reasons of
  parsimony, as it often is (though it \emph{does} have a significant
  empirical burden).
\item
  We demonstrated how we can extract empirical properties predicted by
  operation \(C\) from \emph{Cognitive Mechanics}, including discrete
  logical relations and statistical properties that could be identified
  in material systems.
\end{enumerate}

A suite of several other mental operations similar to operation \(C\) is
explored in \emph{Cognitive Mechanics}, which is now available as a free
downloadable PDF at https://www.cognitivemechanics.org.

\bibliographystyle{apalike}
\bibliography{references}

\end{document}